\documentclass[11pt, a4paper, goog]{google}

\usepackage[authoryear, sort&compress, round]{natbib}
\uselogo{} 

\title{Diffusion Language Models are Super Data Learners}

\author{
Jinjie Ni$^1$\thanks{Correspondence to: Jinjie Ni <\texttt{jinjieni@nus.edu.sg}>}, Qian Liu, Longxu Dou$^2$, Chao Du$^2$, Zili Wang$^3$, Hang Yan$^4$, \authorcr Tianyu Pang$^2$, Michael Qizhe Shieh$^1$\\
$^1$National University of Singapore, $^2$Sea AI Lab, $^3$StepFun, $^4$Shanghai Qiji Zhifeng Co., Ltd.
}

\renewcommand{\today}{}

\begin{document}

\renewcommand{\absfont}{\linespread{1.2}\fontsize{10}{12}\selectfont}

\newlength{\ablabelwidth}
\setlength{\ablabelwidth}{1.8em} 

\newcommand{\abitem}[2][]{%
  \par\noindent
  \hangafter=1
  \hangindent=\ablabelwidth
  \makebox[\ablabelwidth][l]{\textbullet\hspace{0.5em}}
  \if\relax\detokenize{#1}\relax
    #2%
  \else
    {\bfseries #1} #2%
  \fi
}

\definecolor{blue1}{rgb}{0.02352941176,0,1}
\definecolor{jinjie_color}{rgb}{0.43, 0.43, 0.88} 
\newcommand{\jinjie}[1]{\xxcomment{jinjie_color}{J}{N}{#1}}
\newcommand{\xxcomment}[4]{\textcolor{#1}{[$^{\textsc{#2}}_{\textsc{#3}}$ #4]}}
\newcommand{\longxu}[1]{\textcolor{purple}{$_{lx}$[#1]}}

\begin{abstract}
\textbf{Under strictly controlled pre‑training settings, we observe a \textit{Crossover}: when unique data is limited, diffusion language models (DLMs) consistently surpass autoregressive (AR) models by training for more epochs. The crossover shifts later with more or higher‑quality data, earlier with larger models, and persists across dense and sparse architectures. We attribute the gains to three compounding factors: (1) any‑order modeling, (2) super‑dense compute from iterative bidirectional denoising, and (3) built‑in Monte Carlo augmentation; input/parameter noise improves AR under data constraint but cannot close the gap. At scale, a 1.7B DLM trained with a $\sim\!1.5$T‑token compute budget on 10B unique Python tokens overtakes an AR coder trained with strictly matched settings. In addition, a 1B-parameter DLM achieves > 56\% accuracy on HellaSwag and > 33\% on MMLU using only 1B tokens, without any special tricks, just by repeating standard pre-training data. We also show that rising validation cross‑entropy does not imply degraded downstream performance in this regime.}
\end{abstract}

{\renewcommand{\thefootnote}{\textdagger}
\maketitle
}

\renewcommand{\thefootnote}{}
\footnotetext{\textcolor{blue1}{This is an initial draft that will be further improved.}}
\renewcommand{\thefootnote}{\arabic{footnote}}

\section{Introduction}
\label{sec:intro}

Autoregressive (AR) language models have been the default recipe for modern LLMs: causal factorization, teacher forcing delivering high signal-to-FLOPs training and efficient model serving \citep{brown2020language,raffel2020exploring}. Yet the regime in which AR thrived—ever-growing curated corpora paired with limited compute—is shifting. High-quality data is becoming the primary bottleneck, while compute continues to scale \citep{SevillaRoldan2024ComputeGrowth,CommonCrawl2025CrawlSize}. This raises a central question for the next phase of scaling: \emph{which modeling paradigm extracts more signal per unique token when data, not FLOPs, is the constraint?}

We revisit a strong modeling paradigm–diffusion language models (DLMs) with masked/absorbing discrete diffusion \citep{lou2023discrete,shi2024simplified,ou2024your,nie2025large}. DLMs exhibit several modeling advantages over AR models. Such as any-order modeling, on-the-fly context modification, multi-token generation, high compute-to-parameter ratio, etc. In this work, we focus on their data potential, studying DLMs and AR models under a common pre-training regime and ask how far each can go when unique data is scarce but repetitions are allowed.

The main empirical finding is a \emph{Crossover}: when total training tokens are fixed but the number of unique tokens is limited, DLMs consistently surpass equally sized AR counterparts. This crossover is not an isolated artifact—it systematically shifts with core factors. With more unique data, it shifts later; with higher data quality, it shifts later; with larger models, the crossover arrives earlier; and it persists across dense and sparse (MoE) architectures (Figures~\ref{fig:vary_data_quality}, \ref{fig:vary_model_size}, \ref{fig:vary_sparsity}). Under compute-bound settings with abundant unique data, AR recovers its edge by fitting the data more rapidly; but in data-bound regimes, which is our focus and, increasingly, the practical reality, DLM is the final winner.

Why does the crossover occur? Our analysis isolate three contributors. (i) \textbf{Any-order modeling} relaxes AR’s causal inductive bias, increasing the degrees of freedom available to fit the same dataset. (ii) \textbf{Super-density}: DLMs use substantially more FLOPs at train and test time (temporal refinement with bidirectional attention), enabling them to mine limited data more thoroughly and scale the test-time compute. (iii) \textbf{Richer Monte Carlo augmentation}: the diffusion objective explicitly averages over corruption patterns, turning each sequence into many informative variants. Injecting noise into AR inputs or parameters (dropout) does help under data scarcity, but cannot close the gap (Figures~\ref{fig:mask_input_put_together} and \ref{fig:dropout_put_together}). We also clarify a common diagnostics pitfall: rising validation cross-entropy (``overfitting'') does not necessarily imply degraded downstream accuracy (Figures~\ref{fig:ar_loss_eval_trend}, \ref{fig:benchmark_likelihood}). Note that DLMs do overfit with enough epochs and small enough unique data, but significantly later than AR models, and they extract markedly more value before saturation; e.g., a $1$B DLM trained for $480$ epochs on $1$B tokens reaches $\sim56\%$ HellaSwag and $\sim33\%$ MMLU without clear saturation (Figure~\ref{fig:scaled_diffusion_480epochs}). 

To assess the data-efficiency advantages of DLMs in realistic large-scale settings, we trained two 1.7B-parameter models with a total budget of 1.5T tokens' compute, with AR and diffusion objective. Models are trained on 10B unique Python tokens for $\approx$150 epochs, where 10B is an amount representative of practical high-quality data availability for certain programming languages. We observe clear performance crossovers on coding benchmarks, with the resulting diffusion coder achieving parity with state-of-the-art AR code models trained on trillions of unique tokens.

\paragraph{Key takeaways.}
\begin{itemize}
\item \textbf{Intelligence Crossover.} Under limited unique data, DLMs surpass AR models of the same size; the crossover is robust across data budgets, model scales, and sparsity choices. We also ran 2 scaled-up runs up to 1.5 trillion tokens to verify the universal appearance of the crossover phenomenon.
\item \textbf{Data potential.} DLMs achieve roughly $>$3$\times$ data potential compared with AR models, making them strong candidates when data, not FLOPs, is scarce.
\item \textbf{Factors driving the gains.} Any-order modeling + super-dense compute + built-in noisy augmentation are jointly responsible; AR with input/parameter noise helps but cannot match DLMs.
\item \textbf{Scaling trends.} More unique data moves the crossover later; larger models move it earlier; higher data quality moves it later. Sparse AR models perform badly under data constraint, while DLMs benefit consistently from scale regardless of sparsity.
\item \textbf{Caveats.} Rising validation loss need not imply worse downstream performance.
\item \textbf{Compute trade-off.} DLMs require more training and (parallelizable) inference FLOPs to reach their potential; when unique data is abundant, compute-efficient AR can still win.
\end{itemize}

Together, these results sharpen a forecast for the data-bound era: if high-quality tokens remain the scarcest resource, DLMs are a compelling modeling paradigm for pushing frontier capability per unit of unique data, even at the cost of greater compute.

\section{Preliminaries}
\label{sec:preliminaries}

\subsection{Autoregressive Language Models}\label{subsec:ar_preliminary}

Autoregressive (AR) language modeling is the mainstream modeling scheme in state-of-the-art LLMs. It parameterizes the joint distribution over a tokenized sequence $x_{1:T}\in\{0,\dots,K{-}1\}^T$ via the chain rule,
\begin{equation}
    p_\theta(x_{1:T})=\prod_{i=1}^{T} p_\theta\!\left(x_i \mid x_{<i}\right).
    \label{eq:ar_eq1}
\end{equation}
Modern decoder-only LLMs realize this factorization with stacks of causal self-attention blocks \citep{radford2019language}: a triangular (causal) mask limits each position’s receptive field to the prefix $x_{<i}$, allowing rich conditioning within the visible context while preserving left-to-right generation. Training uses teacher forcing—shifting the sequence so that all next-token conditionals are learned in parallel—whereas inference proceeds sequentially with KV-caching for efficiency. This recipe yields exact, normalized likelihoods; supports streaming generation and controllable decoding; and under sufficient scale enables strong in-context learning behaviors \citep{brown2020language}.

\paragraph{Learning objective}
AR training maximizes the log-likelihood (equivalently, minimizes cross-entropy) under the data distribution:
\begin{equation}
    \mathcal{L}(\theta)
    = \mathbb{E}_{x_{1:T}\sim p_{\text{data}}}\!\left[
    \sum_{i=1}^{T} -\log p_\theta\!\left(x_i \mid x_{<i}\right)
    \right].
    \label{eq:ar_eq2}
\end{equation}
This objective computes in a single forward pass by predicting the next token at every position. While AR models are inherently one-directional and can suffer exposure bias from teacher forcing, their simplicity, exact normalization, and scalability make decoder-only Transformers the dominant foundation for high-quality text generation.

\subsection{Masked Diffusion Language Models}\label{subsec:dlm_preliminary}

\textbf{Why masked diffusion?}  
DLMs adopt a noising–denoising framework over sequences. Among their variants, \emph{masked diffusion}—also known as absorbing discrete diffusion, which relies on an absorbing transition kernel—has emerged as the most effective formulation \citep{amin2025masking}. Their bidirectional attention and diffusion objective enable any-order modeling, allowing data to be modeled in arbitrary directions during both training and inference. This property is particularly beneficial for tasks requiring non-causal dependencies and back-and-forth reasoning, such as coding \citep{xie2025dream,wu2025fast}, mathematics \citep{deepmind2025geminiDiffusion}, and report generation \citep{han2025deep}. DLMs' bidirectional attention natively support on-the-fly context modification as new content is generated, a desirable feature in these tasks. Multi-token generation is also naively supported by DLMs, providing the foundation for their bleeding fast decoding. Moreover, DLMs spend more parallelable FLOPs at both the inference and training time, leading to its superior data learning capability and potentially stronger reasoning capabilities.

\paragraph{Forward (corruption) process}
Let $K$ be the vocabulary size, $L$ the sequence length, and $m$ the mask token.
Given a clean sequence $x_0\in\{0,\dots,K{-}1\}^L$, define a monotone diffusion schedule
$\alpha_t\in[0,1]$ with $\alpha_0=1$ and $\alpha_1=0$, where $\alpha_t$ is the probability that a token is \emph{clean} (unmasked) at noise level $t\in[0,1]$.
The forward process independently masks tokens: 
\[
q_{t|0}(x_t\mid x_0)
=\prod_{i=1}^{L} q_{t|0}\!\left(x_t^{(i)}\mid x_0^{(i)}\right),\qquad
q_{t|0}\!\left(x_t^{(i)}\mid x_0^{(i)}\right)
=\begin{cases}
\alpha_t,& x_t^{(i)}=x_0^{(i)},\\[2pt]
1-\alpha_t,& x_t^{(i)}=m~,
\end{cases}
\]
so that the expected unmasked fraction at level $t$ equals $\alpha_t$.

\paragraph{Reverse (denoising) process}
Starting from the fully masked sequence $x_1$ and a decreasing schedule $1=t_0>t_1>\dots>t_N=0$, the reverse dynamics from $t$ to $s<t$ acts independently across positions:
\[
q_{s|t}\!\left(x_s^{(i)} \mid x_t\right)=
\begin{cases}
1, & x_t^{(i)}\neq m,\; x_s^{(i)}=x_t^{(i)},\\[4pt]
\displaystyle\frac{1-\alpha_s}{\,1-\alpha_t\,}, & x_t^{(i)}=m,\; x_s^{(i)}=m,\\[8pt]
\displaystyle\frac{\alpha_s-\alpha_t}{\,1-\alpha_t\,}\;q_{0|t}\!\left(x_s^{(i)}\mid x_t\right), & x_t^{(i)}=m,\; x_s^{(i)}\in\mathcal{V}\setminus\{m\},\\[8pt]
0, & \text{otherwise.}
\end{cases}
\]
i.e., already-revealed tokens stay fixed; masked tokens either remain masked with probability $\frac{1-\alpha_s}{1-\alpha_t}$ or are revealed by sampling from a \emph{data-prediction} distribution $q_{0|t}(\cdot\mid x_t)$ with probability $\frac{\alpha_s-\alpha_t}{1-\alpha_t}$.
A key \emph{time-agnostic} property \citep{ou2024your} of masked diffusion is that
\[
q_{0|t}\!\left(x_0^{(i)}\mid x_t\right)=p_{\text{data}}\!\left(x_0^{(i)} \,\middle|\, x_t^{\text{UM}}\right),
\]
the conditional distribution of the clean token depends only on the \emph{unmasked} context $x_t^{\text{UM}}$; it does not depend on $t$ beyond which tokens are visible. This allows the denoiser to be parameterized without an explicit time embedding.

\paragraph{Learning objective}
Let $p_\theta\!\left(x_0^{(i)}\mid x_t\right)$ approximate $p_{\text{data}}\!\left(x_0^{(i)}\mid x_t^{\text{UM}}\right)$.
Masked diffusion maximizes a variational bound on $\log p_\theta(x_0)$, which can be written as minimizing
\begin{equation}
\mathcal{L} \;=\; \int_{0}^{1} w(t;\alpha)\;
\mathbb{E}_{q_{t|0}(x_t\mid x_0)}\!\left[
\sum_{i:\,x_t^{(i)}=m} -\log p_\theta\!\left(x_0^{(i)}\mid x_t\right)
\right]\mathrm{d}t,
\label{eq:mdm_loss}
\end{equation}
where the importance weight $w(t;\alpha)$ depends only on the schedule and, up to a constant factor, takes the natural form
\[
w(t;\alpha)\;=\;\frac{\alpha'_t}{\alpha_t-1}\,.
\]
Intuitively, $w(t;\alpha)$ compensates for the varying expected number of masked positions across noise levels. For the widely used linear schedule $\alpha_t=1-t$, this reduces to the familiar integrand weight $w(t)=1/t$.

\section{The Intelligence Crossover}
We present extensive results showing that, when trained on standard web tokens, masked DLMs consistently outperform AR counterparts across model sizes in data-constrained settings, achieving higher potential without performance saturation. To analyze this pattern more rigorously, we decompose its key factors and conduct controlled group experiments.

\subsection{Experimental Settings} 
\label{sec:experimental_settings}

Unless specifically mentioned, most of the below sections use the same basic settings. 
It is worth noting that the hyperparameters adopted in our experiments are primarily optimized for AR models, reflecting extensive prior tuning efforts by the broader LLM research community. Although we aimed to maintain identical settings across AR and diffusion models, this is inherently unfair for diffusion models. Consequently, the observed performance advantages of diffusion models could be under-estimate.

All training runs were conducted using a significantly modified Megatron-LM codebase. Cross-over experiments were trained on a subset of the Nemotron-CC corpus \citep{su2024nemotron}; the runs in Figure \ref{fig:combined_loss_vs_epochs} utilized a subset of the c4-en corpus \citep{raffel2020exploring}; the coders are trained on a subset of the RefinedCode \citep{huang2024opencoder}. Note that all token budgets used are randomly sampled from these corpus, without any special process. We used the same masked diffusion objective as in \cite{nie2025large}, detailed in Equation \ref{eq:mdm_loss}. Specifically, we employed a batch size of 256, a sequence length of 2048, and a warmup-stable-decay (WSD) learning rate schedule peaking at 2e-4 with 1000 warmup steps, followed by a 10\% exponential decay to 2e-5. Model parameters were randomly initialized from a normal distribution with s.t.d. 0.02. We adopted a performant architectural configuration, incorporating the GPT-2 tokenizer, RoPE, SwiGLU, pre-layer RMSNorm, bias-free, and qk normalization. All mixture-of-expert models in this work use token-choice routing with 1e-2 auxiliary loss and 1e-3 z loss \citep{zoph2022st}. Validation loss was evaluated on the c4-en validation set using distinct 100M-token subsets per evaluation. Benchmark evaluations strictly adhered to official protocols, detailed in Table \ref{tab:eval_settings}. 

\subsection{Data Budget Decides the Crossover Timing}
\label{subsec:crossovers_across_data_budgets}

\begin{figure}[t!]
\centering
\includegraphics[width=0.8\textwidth]{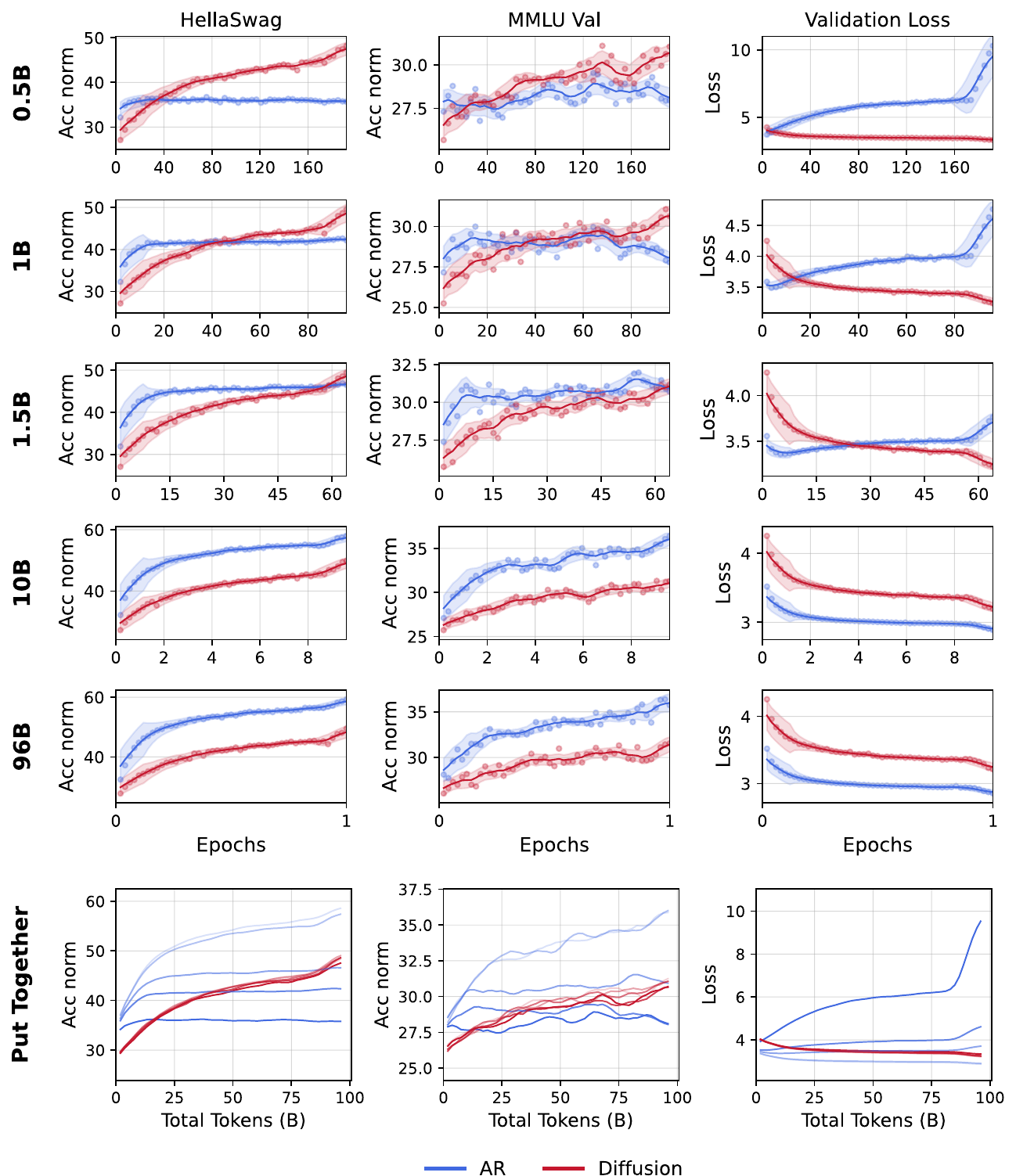}
\caption{\textbf{Diffusion vs. AR with various unique data budgets.} All models are trained on 96B total tokens (including repetition), varying the unique tokens from 0.5B to 96B. In the bottom "Put Together" row, a darker color means a smaller unique data size, i.e., trained for more epochs.}
\label{fig:vary_data_budget}
\end{figure}

The unique data budget is the first key factor we ablate under data-constrained training. Here, we vary the number of unique tokens from 0.5B to 96B while fixing the total training tokens at 96B, constrained by computational resources, and keep the model size fixed at 1B. Notably, 0.5B is a realistic regime, as many domain-specific datasets—such as robotics logs, clinical records, and low-resource language corpora—often fall within this scale. 

As shown in Figure \ref{fig:vary_data_budget}, we consistently observe crossover behavior: diffusion language models surpass their autoregressive counterparts at low data budgets. Our results indicate that DLMs exhibit more than a threefold higher effective data efficiency compared to AR models. Empirically, a DLM trained on only 0.5B unique tokens (not fully converged) achieves comparable performance to an AR model trained on 1.5B unique tokens (converged).

Under compute-bound settings—where data is abundant—AR models fit the training distribution more effectively and achieve stronger end-of-training performance. In contrast, under data-bound conditions—reflecting today’s reality where compute growth outpaces high-quality data availability—diffusion models eventually outperform AR models. Interestingly, the crossover timing is similar across both evaluation benchmarks. As the number of unique tokens increases, the crossover shifts later (0.5–1.5B) or beyond our observable range (1.5–96B). This delay is more clearly highlighted in \S \ref{sec:diffusion_also_overfits}.  

We also aggregated all runs into a single plot for global comparison. Strikingly, DLMs exhibit minimal degradation across both benchmarks and the validation set even when the unique data budget is reduced from 96B to 0.5B tokens—demonstrating substantially higher data efficiency than typically assumed. The narrow gap between the 10B and 96B AR runs reflects the fact that crossover points are pushed further out, meaning that what we observe within the 0–96B range captures only the early phase, where differences remain modest. This also suggests that, despite being more sensitive to data repetition, current AR training pipelines may underutilize available data: repeating the given 10B tokens for 10 epochs leads to only mild short-term degradation, more than the 4 epochs empirically observed in \citet{muennighoff2023scaling}.

\subsection{Varying Data Quality}
\label{subsec:crossovers_across_data_quality}

\begin{figure}[t]
\centering
\includegraphics[width=0.8\textwidth]{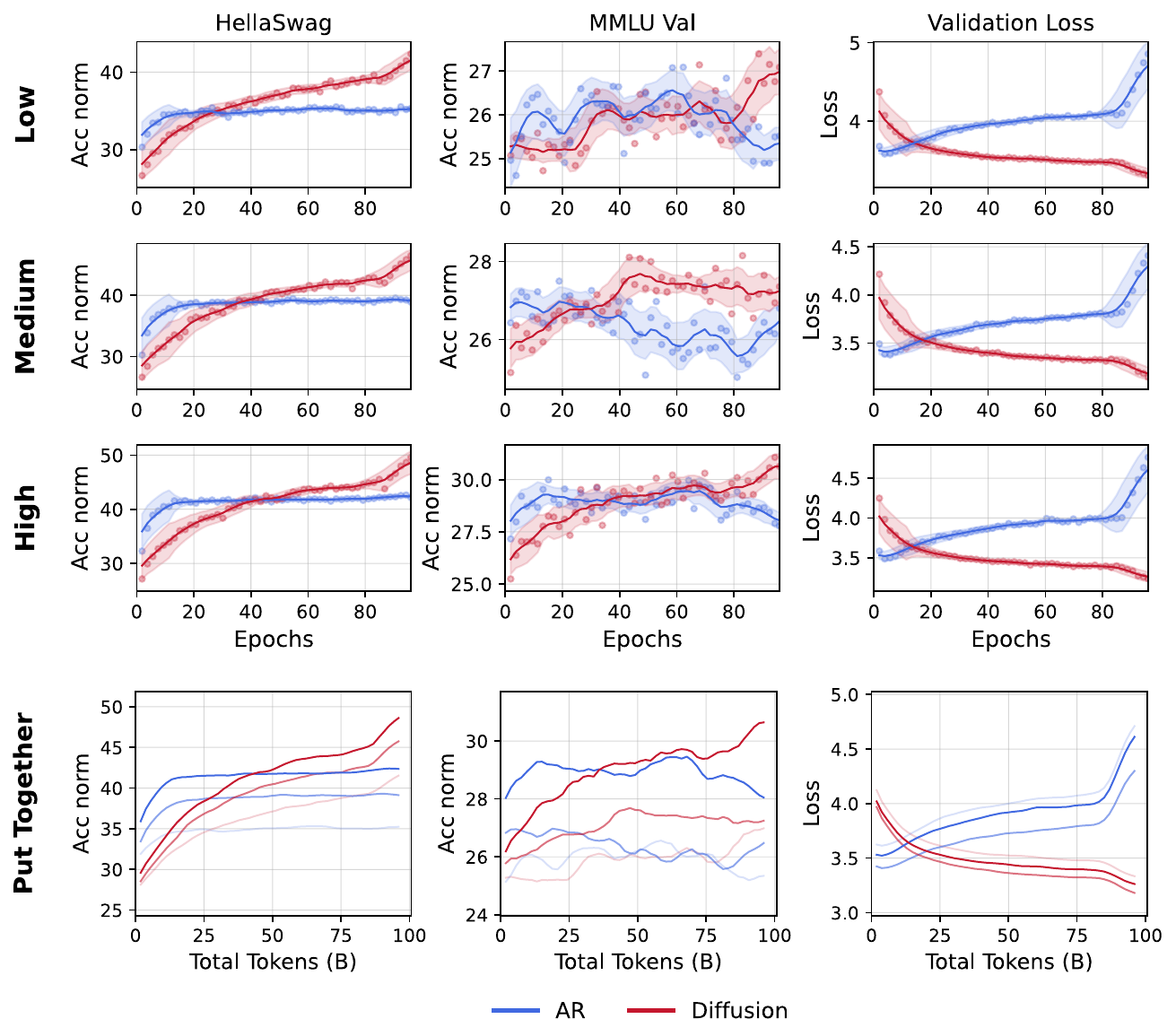}
\caption{\textbf{Diffusion vs. AR with various data qualities.} All models are trained on 1B unique tokens for 96 epochs. In the bottom "Put Together" row, a darker color means a higher data quality.}
\label{fig:vary_data_quality}
\end{figure}

Unique data size is not the only factor influencing scaling behavior; data quality is equally critical. Prior studies show that quality can even outweigh quantity, i.e., "less is more" in LLM training \citep{zhou2023lima,magnusson2025datadecide}. To isolate this effect, we conduct controlled experiments on datasets of varying quality. Specifically, we train 1B-parameter AR and diffusion models on three quality tiers—low, medium, and high—sampled from the same distribution \citep{su2024nemotron}, using 1B unique tokens over 96 epochs.

One might expect the crossover to amplify with higher data quality, since diffusion models typically outperform AR models in data-limited regimes. Our ablation shows that both diffusion and AR models benefit substantially from improved data quality across benchmarks (Figure \ref{fig:vary_data_quality}). However, as quality increases, the crossover shifts slightly later, suggesting that AR models are more sensitive to quality variation. Notably, validation loss trends diverge from benchmark results: the medium-quality run achieves the lowest validation loss. This highlights that validation loss is not always a reliable comparator, as model overconfidence can heavily distort cross-entropy. We expand on this in \S \ref{sec:overfitting_trend}.

\subsection{How Much Does Model Size Impact the Data-Constrained Training?}
\label{subsec:crossovers_across_model_sizes}

\begin{figure}[t]
\centering
\includegraphics[width=0.8\textwidth]{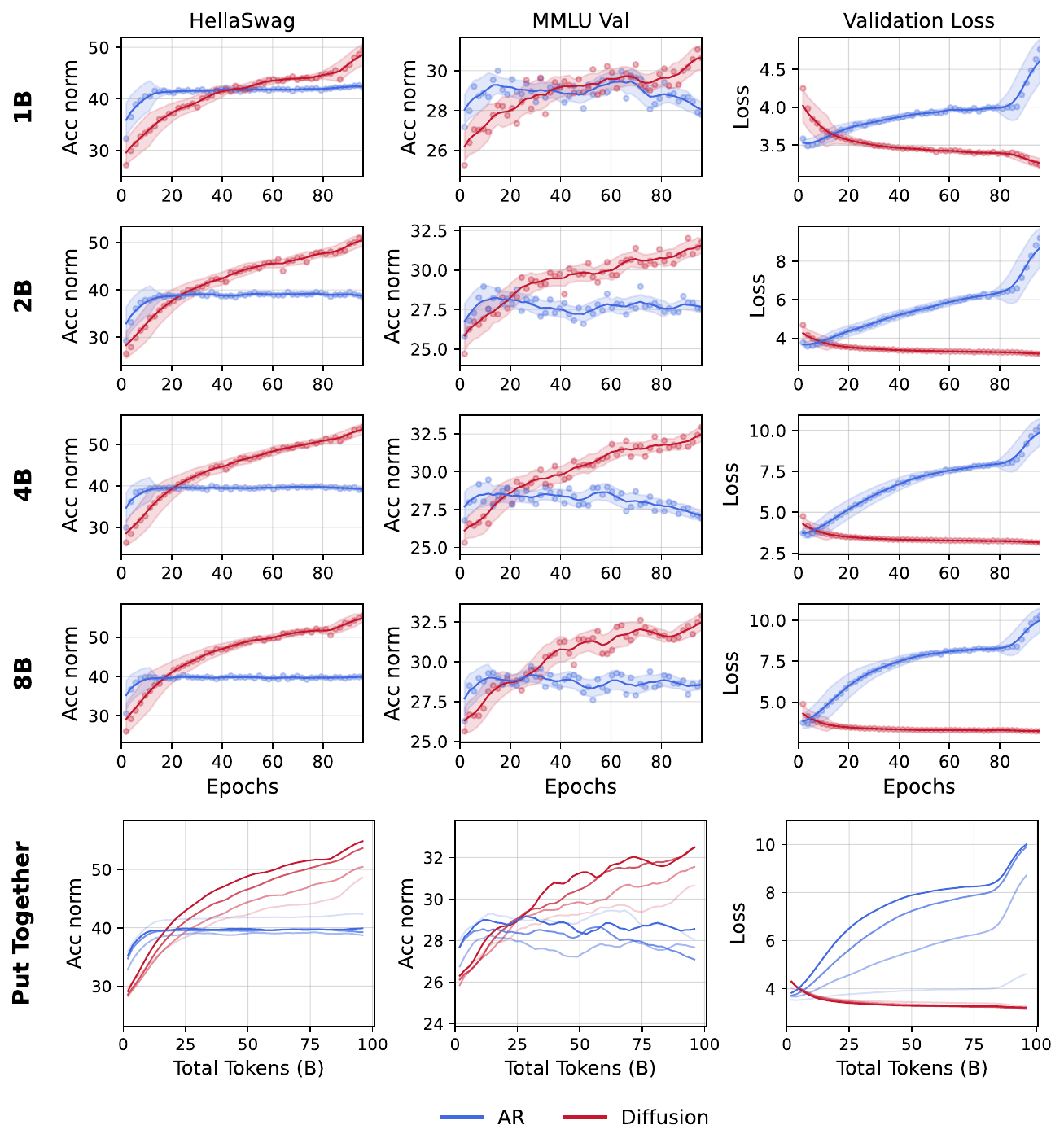}
\caption{\textbf{Diffusion vs. AR with various model sizes ranging from 1B to 8B.} All models are trained on 1B unique tokens for 96 epochs. In the bottom "Put Together" row, a darker color means a larger model size.}
\label{fig:vary_model_size}
\end{figure}

According to compute- and data-constrained scaling laws \citep{hoffmann2022training,muennighoff2023scaling}, model size should grow with the data budget and training epochs. We investigate whether both model families benefit from increased scale and how the crossover evolves. To this end, we train diffusion and AR models from 1B to 8B parameters on 1B unique tokens for 96 epochs, keeping all other settings identical (\S \ref{sec:experimental_settings}).  

As shown in Figure \ref{fig:vary_model_size}, the crossover shifts consistently earlier with larger models. This arises because, under data constraints, AR models quickly saturate the available data, and further scaling not only yields diminishing returns but also increases overfitting (see aggregated plots, bottom). In contrast, diffusion models do not fully exploit the 96B-token budget even at 1B parameters; scaling them accelerates learning and consistently improves performance. Remarkably, under these constraints, even the smallest diffusion model outperforms AR models at all sizes.  

Conceptually, diffusion models learn any-order mappings: an input sequence of length $L$ can be corrupted into $2^L$ variations, whereas AR models only learn causal mappings with $L$ sequence prefixes. This combinatorial expansion makes the diffusion learning space—and thus its upper performance bound—much larger than AR under limited data. While not every corruption yields distinct signal, the enlarged hypothesis space requires larger models to fully capture. We discuss this further in \S \ref{sec:discussions}.

\subsection{Sparse, Dense, Super-Density}
\label{subsec:crossovers_across_sparsity}

\begin{figure}[t]
\centering
\includegraphics[width=0.8\textwidth]{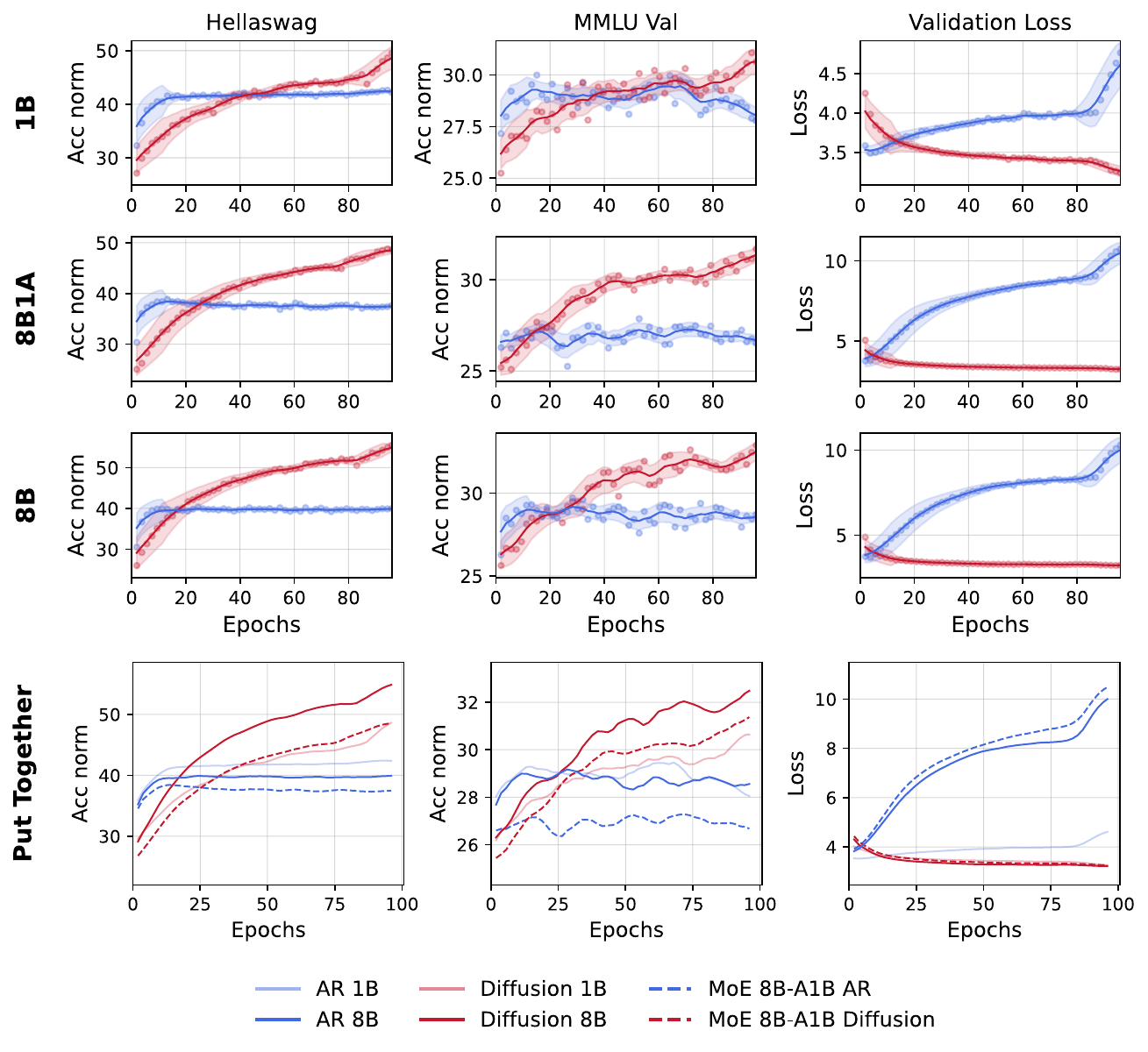}
\caption{\textbf{Diffusion vs. AR with various model sparsities.} All models are trained on 1B unique tokens for 96 epochs. Both 8B-A1B and 8B1A denote 8B total parameters and 1B activated parameters, parameter-matching the 8B dense and FLOPs-matching the 1B dense.}
\label{fig:vary_sparsity}
\end{figure}

A key distinction between AR and diffusion models lies in compute efficiency. Compared to dense AR models, diffusion models require substantially more FLOPs during training to realize the full data potential, and far higher parallel FLOPs during inference to scale along the time axis. We term these diffusion-based dense models "super-dense" architectures. Further analysis of training- and test-time compute trade-offs is provided in \S \ref{sec:discussions}. Building on this, a natural question arises: how do dense and super-dense models compare against sparse architectures, and does sparsity alter performance under data-constrained regimes?

To study the effect of sparsity in data-constrained learning, we train Mixture-of-Experts (MoEs) \citep{shazeer2017outrageously} for both AR and diffusion models, using 8B total parameters with 1B active per forward pass (8B1A). For direct comparison, we also train 1B dense models to match the activated parameters (FLOPs-matching) and 8B dense models to match the total parameters (parameter-matching). The density ordering is: AR MoE < AR dense < DLM MoE < DLM dense. All models are trained on 1B unique tokens for 96 epochs.

Figure \ref{fig:vary_sparsity} reveals several key observations. First, across all sparsity levels, DLMs consistently surpass AR models, with crossover timing ordered as 8B dense < 8B1A MoE < 1B dense. Second, when comparing dense and sparse variants, the evaluation can be framed in two complementary ways: (1) under FLOPs-matching, how much benefit arises solely from increasing total parameters? (2) under parameter-matching, what is the performance cost of reducing per-task FLOPs?

Given the distinct behaviors of AR and diffusion models under data constraints, we analyze them separately.  
For AR models, which fit the data within a few epochs and then saturate (or overfit on the validation set), scaling from 1B to 8B parameters consistently degrades performance across evaluations, regardless of dense or sparse expansion. Sparse expansion (1B dense $\rightarrow$ 8B1A MoE) performs worst. Alternatively, framing the comparison as FLOP reductions from the 8B dense baseline reveals that reducing only FLOPs (8B dense $\rightarrow$ 8B1A MoE) hurts performance, while reducing both FLOPs and parameters (8B dense $\rightarrow$ 1B dense) improves it. Together, these perspectives lead to a clear conclusion: \textbf{in data-constrained settings, given fixed parameter counts, higher FLOPs consistently improve performance}, as evidenced by 8B dense consistently outperforming 8B1A MoE.  

Diffusion models behave differently: within the 96B-token window, they have not yet reached diminishing returns and remain far from overfitting (though \S \ref{sec:diffusion_also_overfits} shows eventual overfitting). In this regime, the 8B1A diffusion MoE performs as expected, generally between the 8B dense and 1B dense counterparts, with its advantage most pronounced on knowledge-heavy benchmarks (e.g., MMLU). We believe the interplay of compute and parameterization under data constraints deserves a dedicated study.

\subsection{Is Noise Deciding the Game?}
\label{subsec:is_noise_dominating_the_game}

In \S \ref{sec:discussions}, we further analyze the factors underlying the advantage of DLMs over AR models. In summary, DLMs differ from AR models in three ways: (1) any-order modeling, (2) higher training- and inference-time FLOPs (“super-density”), and (3) richer Monte Carlo sampling via noisy data augmentation. We argue that noisy augmentation primarily benefits models in data-constrained regimes, whereas the first two constitute fundamental modeling advantages with universal impact. Ablating these factors is non-trivial: introducing any-order modeling into AR while holding other variables fixed is hard, and similarly, scaling AR FLOPs without altering other dynamics is difficult. Moreover, note that masked DLMs can be viewed as any-order models with more practical tractability.

\begin{figure}[ht]
\centering
\includegraphics[width=0.8\textwidth]{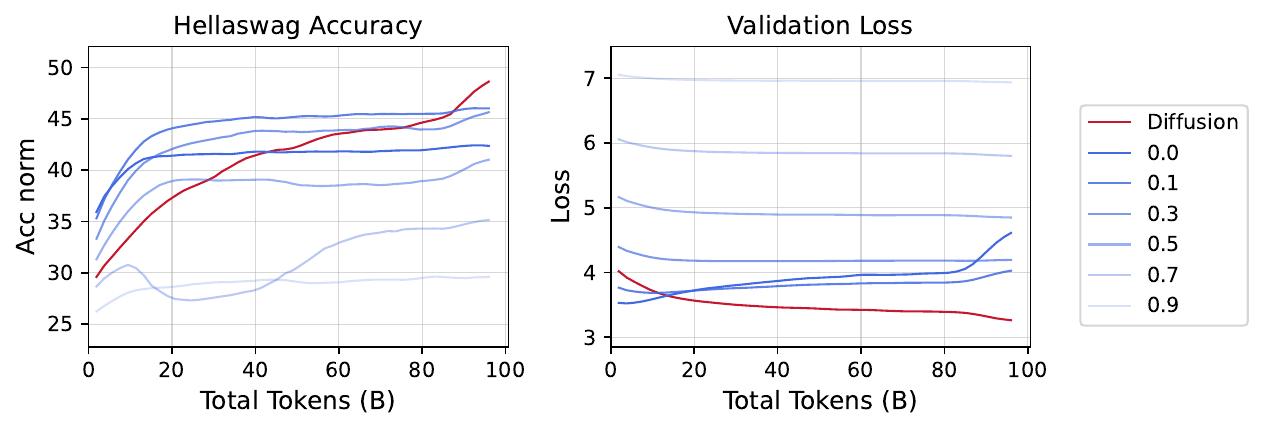}
\caption{\textbf{Injecting noise (random masking) to the AR model's inputs helps in data-constrained settings but does not beat diffusion models.} All models shown in this figure are 1B parameters trained on 1B unique tokens for 96 epochs.}
\label{fig:mask_input_put_together}
\end{figure}

We therefore focus on ablating the data augmentation feature to isolate the contributions of the other two factors. A direct approach is to mimic masked DLMs by injecting noise into AR inputs—masking a fraction of tokens—to evaluate the potential gains. While uncommon in large-scale LLM training, noisy data augmentation has precedent in computer vision. As shown in Figure \ref{fig:mask_input_put_together}, we apply input masking at ratios from 10\% to 90\%. Results show that AR performance improves with modest noise but collapses once inputs become overly corrupted. The substantial gains under low-noise regimes confirm that noisy data augmentation is indeed a key contributor to diffusion’s advantage in data-constrained settings. However, even the best configuration (10\% masking) falls well short of diffusion, and saturates by the end of the 96B-token training window, whereas diffusion remains far from saturation and would likely widen the gap with longer training.

\begin{figure}[ht]
\centering
\includegraphics[width=0.8\textwidth]{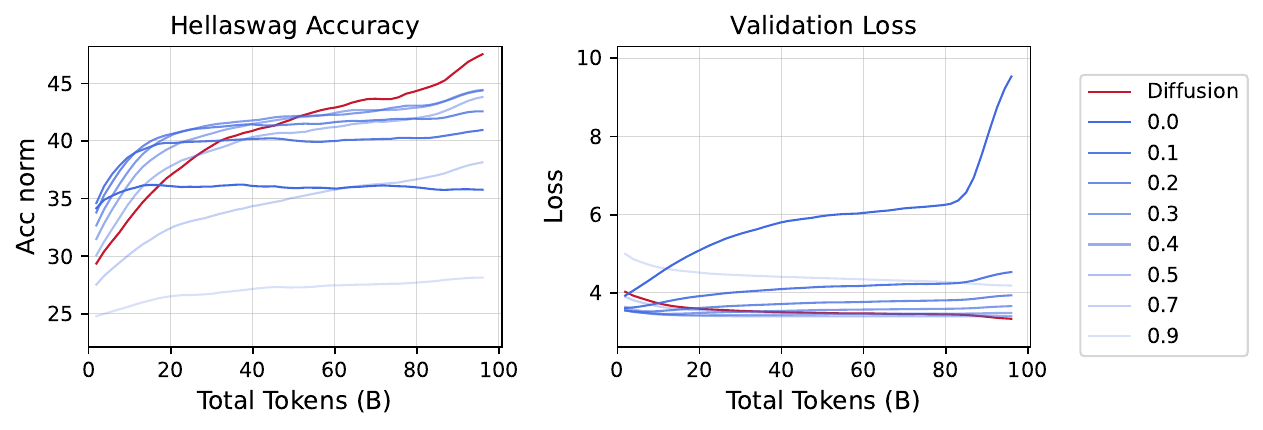}
\caption{\textbf{Injecting noise to AR model's parameters (dropout) helps in data-constrained settings but does not beat diffusion models.} All models shown in this figure are 1B parameters trained on 1B unique tokens for 96 epochs.}
\label{fig:dropout_put_together}
\end{figure}

Alternative to injecting noise to the input for noisy data augmentation, we can also inject noise to the parameters by zeroing out a random set of neuron outputs, a.k.a., dropout, to achieve similar effects. Similarly, we use a dropout ratio from 10\% to 90\%. As shown in Figure \ref{fig:dropout_put_together}, adding noise to the parameter also raises the performance of AR models in data-constrained settings while not fully eliminating the advantage of DLMs. The AR diminishes likewise but diffusion doesn't.

\section{Scaling Crossovers to Trillion-Level Total Tokens}
\label{sec:a_practical_case}

\begin{figure}[ht]
\centering
\includegraphics[width=1.0\textwidth]{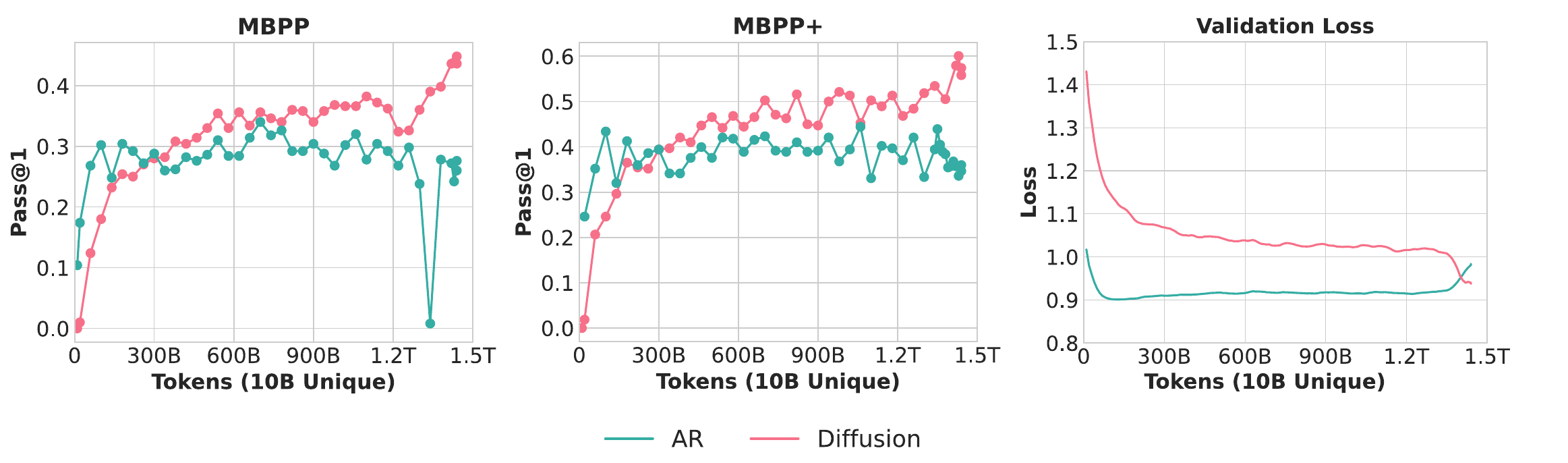}
\caption{\textbf{1.7B AR vs. DLM trained on 10B unique code tokens for $\approx$150 epochs.} On downstream evaluations, we observe early crossovers where DLMs surpass AR models. This provides another evidence that crossovers emerge at larger scales under limited unique data. Different crossover timings are observed on two additional coding benchmarks (Figure \ref{fig:coder_humaneval}).}
\label{fig:coder_mbpp_valloss}
\end{figure}

We further validate the crossover phenomenon under larger unique token budgets and on generative tasks. Among such tasks, code generation provides a particularly clean and popular benchmark, while also being heavily constrained by available training tokens. Prior work suggests that a 10B-token budget is already a practical scale for certain programming languages \citep{huang2024opencoder}. To this end, we construct a 10B-token dataset by randomly sampling from RefineCode \citep{huang2024opencoder}, consisting of 9B unique Python tokens and 1B annealing tokens, ensuring a strictly non-repetitive training corpus. On this dataset, we train a pair of 1.7B-parameter AR and DLM models under identical settings for $\approx$150 epochs. The training setup employs a warmup–stable–decay learning rate schedule (2000 warmup steps, followed by decay over 100B tokens), resulting in a total compute budget of $\approx$1.5T tokens. Additional hyperparameters include a sequence length of 4096, a global batch size of 1024, weight decay of 0.1, and a performant architecture: GPT-NeoX \citep{black2022gpt} tokenizer, rotary position embeddings (RoPE), SwiGLU activations, pre-layer RMSNorm, bias-free layers, and $qk$ normalization.

Interestingly, we observe a very clear crossover in the early stages of training across downstream benchmarks, further reinforcing the apparent universality of the diffusion–AR crossover phenomenon. At the same time, we note that the DLM had not yet converged by the end of the 1.5T-token training cycle, suggesting substantial untapped potential if training were to continue. Beyond standard benchmarks, we evaluate both models on HumanEval and HumanEval+, and find that the crossover occurs at a different point compared to MBPP and MBPP+. Specifically, in HumanEval(-plus), the performance curves intersect only near the end of the annealing stage (see Figure \ref{fig:coder_humaneval}), without seeing diminish returns. We hypothesize that this timing discrepancy is rooted in the evaluation protocols: MBPP and MBPP+ adopt a 3-shot setting, while HumanEval and HumanEval+ use a 0-shot setting. Such differences in evaluation configuration likely amplify or suppress model capabilities at different stages of training, thereby shifting crossover points.

These results highlight two key insights. First, the crossover phenomenon extends robustly to generative tasks, under more unique token budgets. Second, the precise timing of crossovers in generative tasks may be sensitive to evaluation protocols, which raises an important methodological consideration: the need for systematic studies that disentangle training dynamics from evaluation artifacts. We believe a more rigorous examination of evaluation settings will be critical for fully characterizing crossover behavior in future work.

\section{High Validation Loss $\neq$ Degraded Intelligence}
\label{sec:overfitting_trend}

\begin{figure}[h]
\centering
\includegraphics[width=0.9\textwidth]{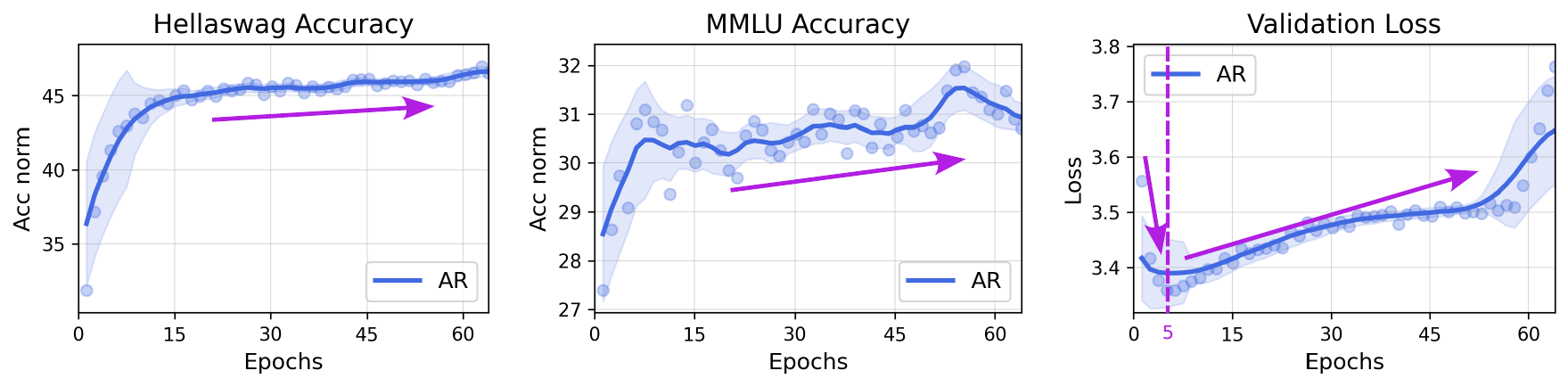}
\caption{When models get “overfit” on pre-training validation sets, their performance on down-stream evaluations doesn’t necessarily drop, and may keep improving till the end of training.}
\label{fig:ar_loss_eval_trend}
\end{figure}

We observe that the AR models exhibiting signs of "overfitting"—indicated by an increase in validation loss—continue to improve on downstream tasks, as illustrated in Figures \ref{fig:ar_loss_eval_trend} and the previous ones. This phenomenon arises because validation loss is measured as an absolute Cross-Entropy loss (Negative Log-Likelihood, NLL), whereas accuracy on multiple-choice benchmarks depends on comparing the relative Cross-Entropy losses across options. Consequently, changes in absolute NLL values do not necessarily translate into changes in their relative ordering.

\begin{figure}[ht]
\centering
\includegraphics[width=0.9\textwidth]{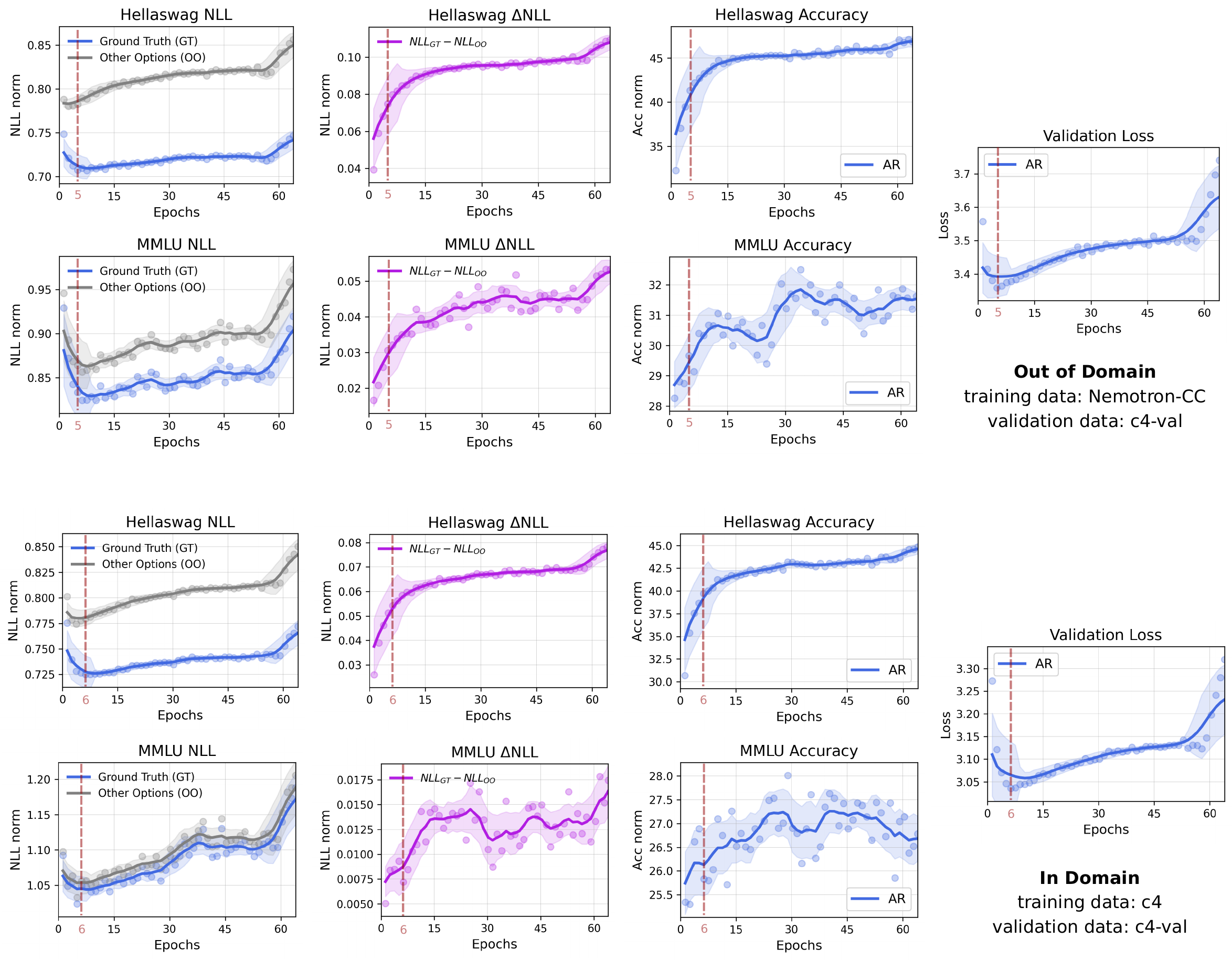}
\caption{\textbf{An illustration of why the models’ performance keeps growing after they overfit on pre-training validation sets (indicated with dashed line)}. NLL: Negative log-likelihood on the ground-truth and other options of multiple-choice evals (NLLs on other options are averaged). $\Delta$NLL: The differences between the NLLs on ground-truth and other options, which keeps growing. This is a 1B autoregressive model trained on 1.5B unique tokens, 64 epochs, on both out-of-domain and in-domain pre-training data.}
\label{fig:benchmark_likelihood}
\end{figure}

In Figure \ref{fig:benchmark_likelihood}, we visualize the average negative log-likelihood (NLL) for the ground-truth and alternative options across multiple-choice evaluations, along with their respective differences ($\Delta$NLL), during the pre-training of a 1B-parameter autoregressive model over 1.5B unique tokens for 64 epochs. Notably, even at the first validation checkpoint (after 3,600 training steps), the model already exhibits substantially lower NLL (higher likelihood) on the ground-truth options, indicating an early capacity to preferentially assign higher logits to correct choices. As training continues, the model begins to overfit, causing an increase in NLL values for both ground-truth and incorrect options. Interestingly, even after this "overfitting," the gap between ground-truth and alternative NLLs continues to widen consistently, indicating that the model's underlying discriminative ability continues to improve despite the rise in validation loss. This phenomenon persists for both in-domain and out-of-domain training data.

One plausible explanation is that repeated exposure to a limited set of training data causes the model to become excessively confident on certain text segments, amplifying NLL values for incorrect predictions. Nevertheless, the persistent growth in relative NLL differences between ground-truth and other options reflects continuous improvement in the model’s discriminative power. A similar rationale applies to generative evaluations, where choices are made at the token rather than sentence level, and we hypothesize that being mistakenly overconfident on non-essential tokens has limited impact on the overall task. The same overfitting patterns on generative benchmarks is revealed in \S \ref{sec:a_practical_case} on code generation benchmarks, where the AR models overfit early on validation loss while the benchmark degradation is typically delayed.

\section{Diffusion Language Models also Overfit the Data}
\label{sec:diffusion_also_overfits}

\begin{figure}[ht]
\centering
\includegraphics[width=0.9\textwidth]{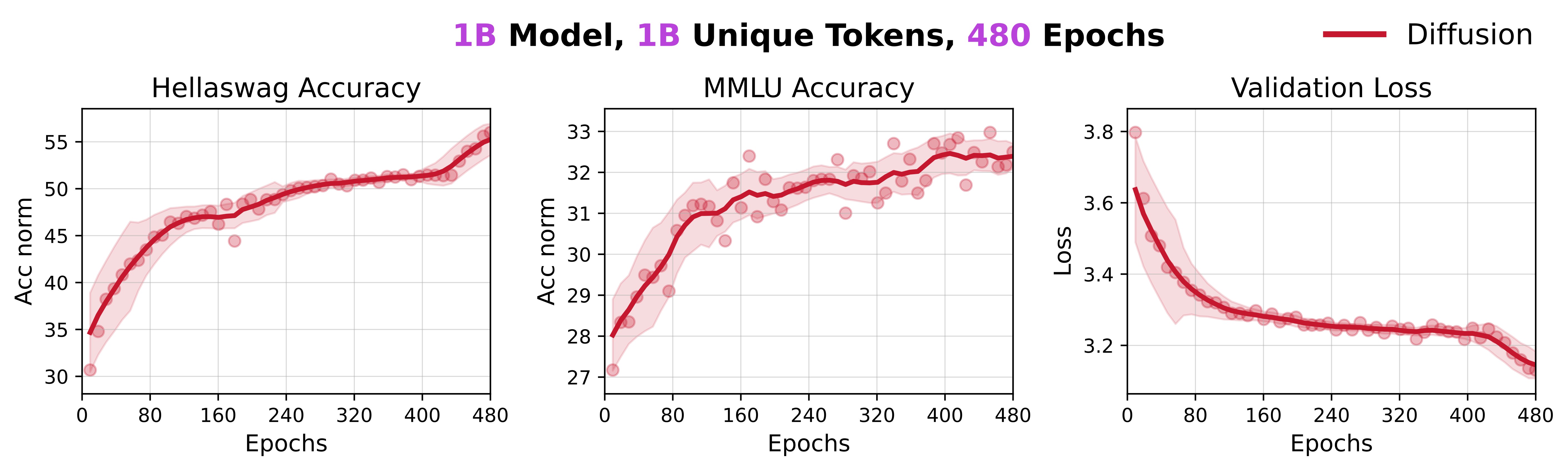}
\caption{The 1B-parameter DLM—trained solely on the original 1B pre-training tokens for 480 epochs—achieves ~56\% accuracy on HellaSwag and ~33\% on MMLU.}
\label{fig:scaled_diffusion_480epochs}
\end{figure}

To study the full potential of tokens in DLM training, we launched an additional run in which the same 1B-token dataset was repeated for 480 epochs, yielding a total of 480B training tokens. Notably, it achieves ~56\% accuracy on HellaSwag and ~33\% on MMLU, significantly outperforming AR’s ~41\% and ~29\%, respectively. Surprisingly, even under such extreme repetition, performance did not saturate, suggesting that DLMs can extract substantially more signal from a fixed 1B-token corpus.  This leads us to investigate: Do DLMs eventually overfit given sufficient training?

\begin{figure}[ht]
\centering
\includegraphics[width=0.9\textwidth]{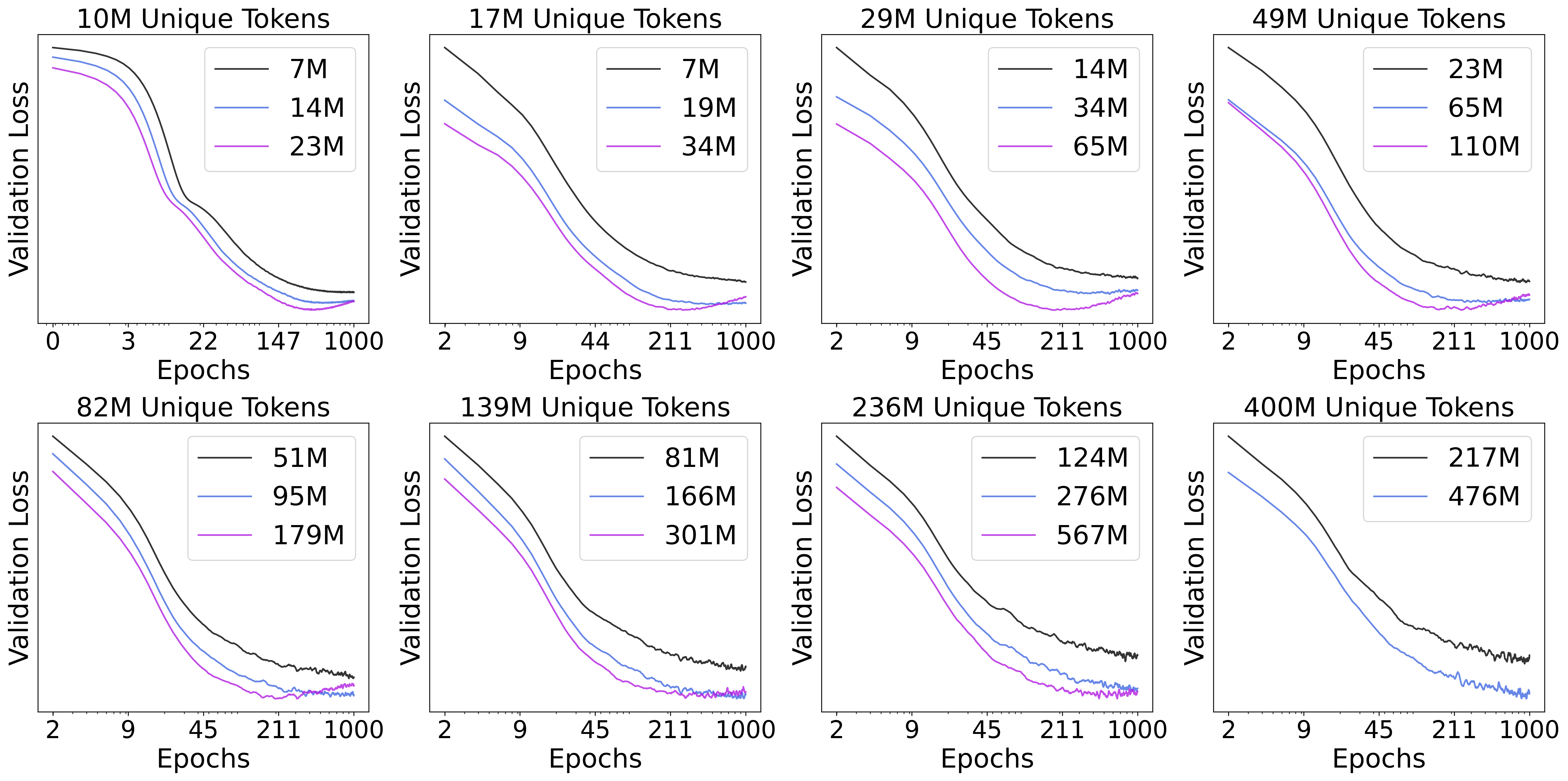}
\caption{\textbf{Validation loss curves for models trained with various sizes and unique data budgets, repeating up to 1000 epochs}. Diffusion language models will also overfit. The more unique data we train on, the later it overfits; the larger model we train, the earlier it overfits.}
\label{fig:combined_loss_vs_epochs}
\end{figure}

We trained models across various sizes and unique data budgets, extending training up to 1000 epochs. As illustrated in Figure \ref{fig:combined_loss_vs_epochs}, when the unique data is sufficiently small and the model size sufficiently large, overfitting eventually emerges after prolonged training.
Specifically, we observe that the epoch at which a model begins to overfit positively correlates with the unique data size and negatively correlates with model size. In other words, larger unique data size delay overfitting, while larger models accelerate its onset.
It is important to note that validation loss overfitting does not immediately imply a decline in model capability—actual performance degradation typically occurs much later (e.g., as seen in Figure \ref{fig:vary_data_budget} at 0.5B tokens and 192 epochs).

\section{Discussions}
\label{sec:discussions}

\subsection{What is the Real Advantage of Diffusion Language Models?}

\paragraph{Reduced inductive bias via any-order modeling} Autoregressive language models impose a strict causal inductive bias on textual data modeling, where each token prediction is conditioned solely on preceding tokens. While natural language exhibits inherent left-to-right causality from a human perspective, evidence indicates that modeling language in reverse or arbitrary order remains feasible \citep{xue2025any}. Moreover, numerous non-causal data types, such as source code, database entries, symbolic notations, and biological sequences, etc., frequently appear online. Thus, enforcing a purely causal inductive bias significantly restricts capturing the rich patterns embedded in diverse textual distributions. DLMs remove this inductive bias with the diffusion objective and the bi-directional attention, enabling such any-order modeling, fully squeezing the value of every single data point.

\paragraph{Super-Density: more training and test time FLOPs per task} As illustrated above, diffusion models outperform AR counterparts by repeatedly processing some portion of unique data during training, effectively scaling FLOPs along the temporal dimension. The continuous-time objective utilized by masked language models is particularly advantageous, enabling high granularity in temporal FLOPs scaling. Similarly, at inference, diffusion models iteratively refine predictions, further amplifying computational density per task. Notably, bidirectional attention implies each token is computed up to N times to generate a sequence of length N, contrasting with AR models using KV cache, which compute each token only once. The effectiveness of super-density in data-constrained settings has been validated in the strictly controlled experiments in \S \ref{subsec:crossovers_across_sparsity}.

\begin{figure}[ht]
\centering
\includegraphics[width=0.9\textwidth]{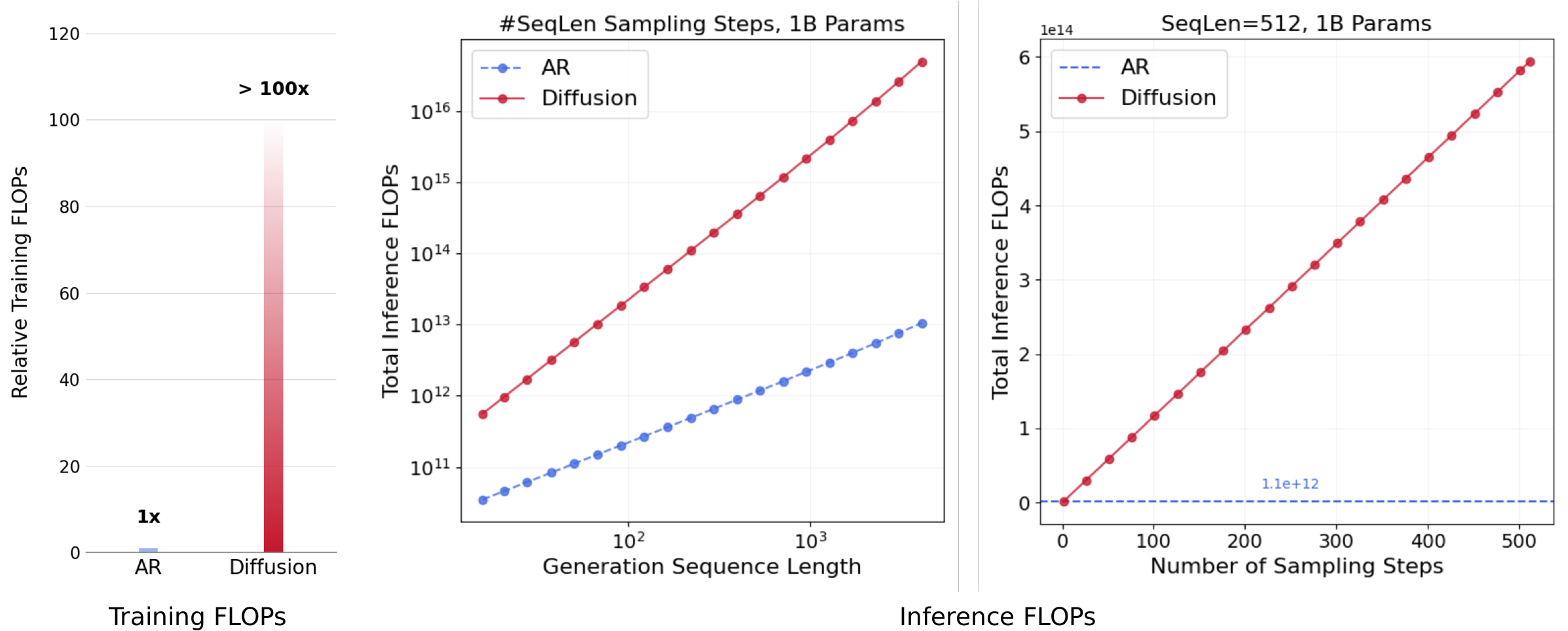}
\caption{(\textbf{left}) The diffusion language models are approximated to consume >100 time more FLOPs than AR counterparts to achieve their full potential in training (where the peak performance is usually much greater than AR). (\textbf{middle}) The theoretical inference FLOPs controlling the sampling steps to be equal to the sequence length. The total inference FLOPs have a power-law relationship with the generation sequence length for both. (\textbf{right}) The theoretical inference FLOPs controlling the generation sequence length, where sampling 512 steps from an AR model with KV cache $\approx$ sampling 1 step from the masked diffusion model.}
\label{fig:train_and_inference_flops}
\end{figure}

Figure \ref{fig:train_and_inference_flops} compares the FLOPs of autoregressive (AR) and masked diffusion models during training and inference. At training time, our preliminary experiments indicate diffusion models require at least about two orders of magnitude (>100×) more FLOPs than AR models to reach optimal performance, with the exact figure varying depending on model size, data budget, etc. During inference, given a fixed number of sampling steps, masked diffusion models consume between 16× and 4700× more FLOPs per task, with this gap widening as the target sequence length increases from 16 to 4096 (Figure \ref{fig:train_and_inference_flops} middle). Moreover, for a constant sequence length, FLOPs consumed by diffusion models scale linearly with the sampling steps. In theory, diffusion models can generate an N-token sequence within a single step, whereas AR models inherently require N sequential steps. However, due to the KV-cache mechanism, the computational cost for AR models generating N tokens is roughly equivalent to that of diffusion models performing a single sampling step. It's worth noting that a significant portion of diffusion model computations can be parallelized. Thus, in practice, before GPU compute bound is reached, the inference speed gap between diffusion and AR models at the same number of sampling steps remains acceptable. Additionally, advances in GPU architectures specifically optimized for compute-intensive workloads may further mitigate this performance gap in the near future. Reflecting on the LLM history, many recent intelligence leaps, such as T5 \citep{raffel2020exploring}, GPT-3 \citep{brown2020language}, and o1 \citep{openai2024learning}, are the direct results of FLOPs scaling.

\begin{equation}
    \mathcal{L} \triangleq  \textcolor{blue}{\mathbb{E}_{t, q(\boldsymbol{x}_t|\boldsymbol{x}_0)}} \left[ \frac{\alpha'_t}{1-\alpha_t} \sum_{\{i | \boldsymbol{x}_t^i = m\}} \log p_{\theta}(\boldsymbol{x}_0^i \mid \boldsymbol{x}_t) \right]
\end{equation}

\paragraph{Learning From Richer Monte Carlo Sampling}
When conducting multi-epoch training for masked DLMs, we effectively transform each unique data point into multiple noisy variants. Specifically, the loss function for masked diffusion models includes an expectation term $\mathbb{E}_{t, q(\boldsymbol{x}_t|\boldsymbol{x}_0)}$, placed outside the negative log-likelihood component. Here, $q(\boldsymbol{x}_t|\boldsymbol{x}_0)$ represents the distribution of masked sequences $\boldsymbol{x}_t$ conditioned on the clean input $\boldsymbol{x}_0$ at diffusion timestep $t$, as determined by the forward corruption process. Intuitively, this means averaging the loss across all possible masking configurations $\boldsymbol{x}_t$ at each time step $t \in [0, 1]$.

In other words, the objective function explicitly requires each data point in the pre-training dataset to be corrupted at multiple masking ratios and combinations for more effective training, by estimating a more precise expectation. Thus, data repetition emerges inherently from the diffusion model's objective rather than from an arbitrary source. Open-source models, such as LLaDA, typically corrupt each data point only once, likely due to computational limitations, approximating the expectation term using a single-sample Monte Carlo estimator. 

In general, masked DLMs corrupt an input data with length L into $2^L$ different sequences, while AR learns causal mapping with $L$ different sequences from short to long for a given input of the same length. Such drastic expansion of the learning space makes DLM's learning upper limit much higher than AR given a limited portion of data (though not every corruption produces a significant difference), requiring larger model size to fully fit it. There have been a lot of frontier works trying to augmenting the AR models' data with model rewriting–rewriting an input sequence into multiple variants, which also demonstrated fruitful gains for AR models \citep{team2025kimi}. However, we note that such data rewriting requires expert design of the whole complicated pipeline and even though it's improving the models' performance, there's not guarantee that the improvement is well-rounded. In contrast, for diffusion language models, injecting noise to the input is all you need to do the data augmentation.

\subsection{Autoregressive Models are Trading Data Potential for Compute Efficiency}

The autoregressive (AR) modeling methodology (decoder-only transformer architecture with teacher-forcing and causal masking) is a legendary local optimum in the AI history. Its success can be broken down into two factors:

\paragraph{Optimal utilization of modern GPU architectures} AR achieves an exceptionally high signal-to-FLOPs ratio during training and a high Model FLOPs Utilization (MFU) during batched inference. During training, each token in a batch consistently receives gradient signals, approximately twice the expected signals compared to masked diffusion models with linear schedules. Indeed, it is challenging to identify alternative methodologies surpassing AR in terms of signal-to-FLOPs efficiency. At inference, token-by-token generation naturally facilitates throughput optimization techniques such as continuous batching, maximizing MFU and KV cache, minimizing computation. Thus, AR stands as an exceedingly robust and efficient baseline method.
\paragraph{Natural language can be causally modeled with low loss} Empirically, pre-training on web-scale corpora demonstrates that left-to-right modeling consistently attains lower loss compared to alternative sequence orders (see Figure 2 of \cite{xue2025any}). If one must select a single sequence order for language modeling, the left-to-right order is empirically optimal (Eq. \ref{eq:ar_eq1}), as it effectively captures natural language patterns. It’s also easy to interpret this: most text data are generated by humans, and humans are RNNs. However, as previously discussed, purely left-to-right modeling inherently misses certain contextual dependencies, indicating room for improvement in data potential.

Currently, there is an emerging trend in which computational resources are becoming increasingly affordable, shifting the primary constraint for scaling intelligence towards data availability. Consequently, for researchers targeting advanced intelligence, the previous emphasis on maximizing GPU utilization has diminished. The inherent modeling limitations imposed by causal masking are now becoming unacceptable. This motivates our exploration of DLMs, which intentionally sacrifice computational efficiency to achieve higher data efficiency—representing an approach diametrically opposed to autoregressive methods.

To strike a favorable balance between these two extremes, a natural strategy is interpolation, as exemplified by block diffusion methods \citep{arriola2025block}. However, achieving comparable training efficiency remains challenging: block diffusion inherently conditions each generated block on a clean context, significantly constraining training efficiency compared to the highly efficient teacher-forcing paradigm employed in autoregressive training.

\subsection{Limitations}
\label{subsec:limitations}
DLMs trade compute for data potential. The crossovers rely on higher train- and test-time FLOPs with bidirectional attention and iterative refinement, which raises energy cost, processing time, and memory pressure compared to AR with KV cache. Whereas under the multi-token prediction scheme it can achieve much lower inference latency than AR models. Second, practical inference performance is sensitive to choices that are still under-explored at scale—masking schedules, denoising weights, step counts, and decoding policies—which complicates fair, apples-to-apples comparisons with heavily optimized AR pipelines. Third, evaluation confounds remain: diffusion objectives optimize a variational bound rather than a normalized left-to-right likelihood, so perplexity is not directly comparable, and benchmark gains may not uniformly translate to streaming, tool-use, or long-horizon generation. Fourth, heavy data reuse heightens contamination and memorization risk if deduplication and auditing are imperfect; safety and privacy audits should therefore be stricter under super-dense training. Finally, the systems stack for practical deployment is less mature for DLMs than for AR, and our experiments focus on English-centric pretraining; broader multilingual, multimodal, and long-context regimes merit dedicated study.

\section{Related Work}
\label{sec:related_work}

\paragraph{Diffusion language models} Building on the theoretical foundations of DLMs \citep{lou2023discrete,shi2024simplified,ou2024your,sahoo2024simple}, \citet{nie2025large} trained the first large-scale DLM from scratch, achieving performance competitive with leading open-source AR models \citep{dubey2024llama}. In parallel, several commercial DLMs have emerged, demonstrating strong coding and math capabilities while offering significantly lower generation latency \citep{deepmind2025geminiDiffusion,khanna2025mercury,song2025seed}.
Efforts have also explored hybrid approaches bridging AR and diffusion. Block diffusion \citep{arriola2025block} performs block-wise diffusion, with block size 1 reducing to AR modeling without shift. Dream \citep{ye2025dream} initialized DLMs with AR priors and employed a "shift-by-one" diffusion strategy to better retain AR knowledge, offering another effective training paradigm. Recent work has also advanced DLM coders \citep{gong2025diffucoder,xie2025dream}, DLM RL scaling \citep{zhu2025llada}, accelerated inference techniques \citep{wu2025fast}, pushing DLMs toward greater practicality and competitiveness.

\paragraph{Mitigating data constraints and the ``token crisis.''}
A complementary line of work asks how to keep scaling when unique tokens are scarce. \citet{muennighoff2025sdc} formalize data-constrained scaling laws, showing that repeating data for up to $\sim$4 epochs yields little loss penalty at fixed compute but exhibits sharply diminishing returns thereafter; they also find that mixing in code or relaxing aggressive filters can substitute for some unique text. \citet{xue2023tokencrisis} analyze multi-epoch degradation, identifying dropout as a robust regularizer; this echoes earlier evidence that small repeated fractions can disproportionately harm generalization \citep{hernandez2022repeat}. In parallel, large open corpora aim to expand high-quality supply through better curation and deduplication: RefinedWeb (5T tokens) \citep{penedo2023refinedweb}, FineWeb (15T; with FineWeb‑Edu) \citep{fineweb2024}, and Dolma (3T) \citep{soldaini2024dolma}.

A second strategy manufactures or selects higher-utility tokens. Rephrasing/rewriting pipelines improve pretraining efficiency by paraphrasing web documents (WRAP) \citep{maini2024wrap}, extending to multilingual rephrasing \citep{pieler2024rephrase} and targeted math/code rewriting (SwallowCode/SwallowMath) \citep{fujii2025rewrite}. Large applied systems report related practices at scale; for example, Kimi K2 describes large agentic data synthesis, and independent commentary attributes part of its gains to rephrasing public corpora \citep{kimi2025k2,breunig2025kimi}. Orthogonal levers include retrieval‑time scaling with trillion‑token datastores \citep{shao2024trillion}, perplexity‑based data pruning \citep{ankner2024perplexprune}, and end‑of‑training domain upsampling \citep{blakeney2024sparkjoy}. Our results are complementary: diffusion objectives further amplify value per unique token when repetition is unavoidable.

\section{Acknowledgment}
We thank Shen Nie, Jiacheng Ye, and Cunxiao Du for their fruitful discussions and pointers.


\bibliography{main}

\appendix

\begin{table}[ht]
\centering
\caption{Summary of downstream evaluation used in this work. CF stands for Completion/Cloze formulation, Gen stands for generative, Char stands for per-character normalization. MMLU combines 0-5 shot results for stability \citep{muennighoff2024olmoe}.}
\label{tab:eval_settings}
\begin{tabular}{llllll}
\toprule
Dataset    & \multicolumn{1}{c}{Format} & \multicolumn{1}{c}{Shot} & \multicolumn{1}{c}{Norm} & \multicolumn{1}{c}{Split} & \multicolumn{1}{c}{Codebase} \\ \midrule
HellaSwag  & CF                         & 0                        & Char                     & Val                       & Megatron LM                  \\
MMLU       & CF                         & 0-5                      & Char                     & Val                       & Megatron LM                  \\
HumanEval  & Gen                        & 0                        & -                        & Test                      & lm-evaluation-harness        \\
MBPP       & Gen                        & 3                        & -                        & Test                      & lm-evaluation-harness        \\
HumanEval+ & Gen                        & 0                        & -                        & Test                      & lm-evaluation-harness        \\
MBPP+      & Gen                        & 3                        & -                        & Test                      & lm-evaluation-harness        \\ \bottomrule
\end{tabular}
\end{table}

\begin{figure}[ht]
\centering
\includegraphics[width=0.7\textwidth]{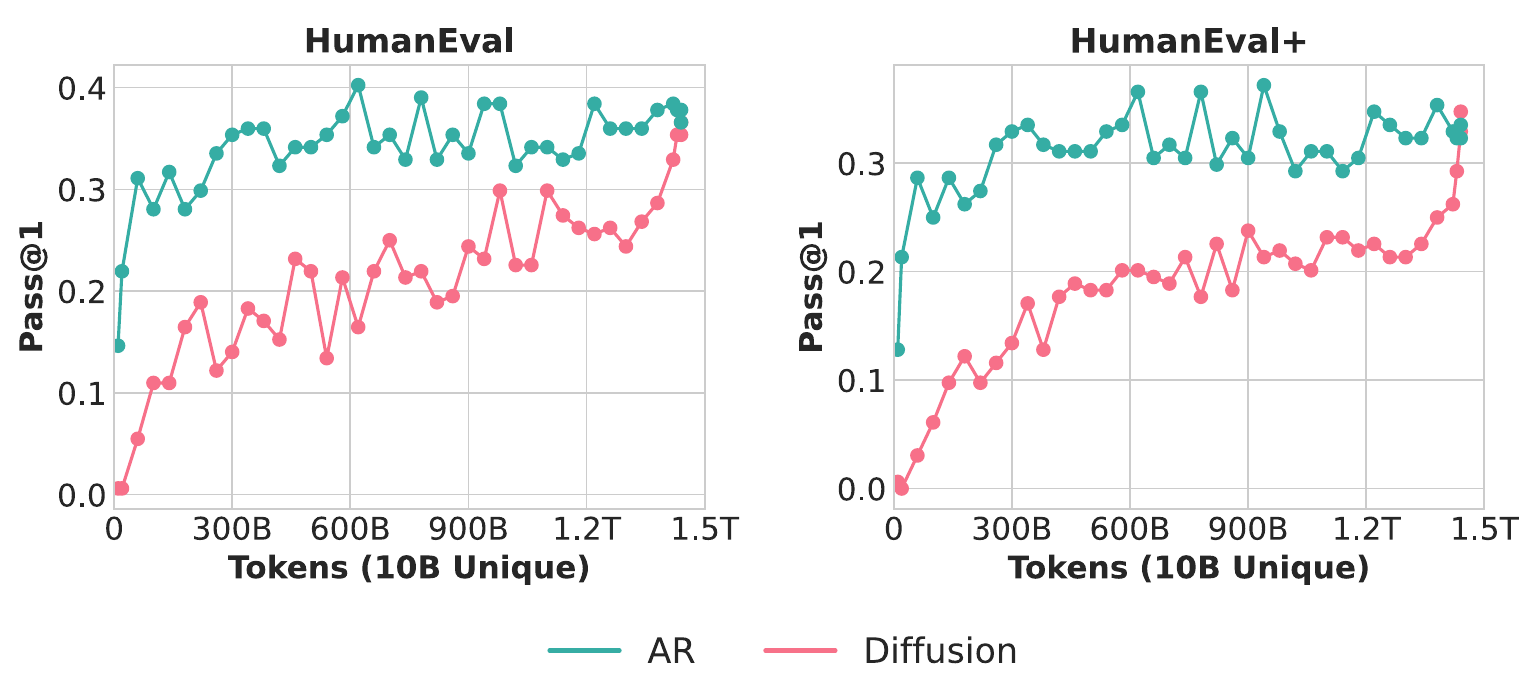}
\caption{\textbf{1.7B AR vs. DLM trained 10B unique code tokens for $\approx$ 150 epochs.} On HumanEval and HumanEval +, the crossover time is delayed and the two curves meet in the end, where the DLM didn't see diminish return. We suspect this is a direct result of the zero-shot setting.}
\label{fig:coder_humaneval}
\end{figure}

\end{document}